\documentclass[runningheads]{llncs}
\usepackage[T1]{fontenc}
\usepackage{graphicx}
\usepackage[colorlinks=true,allcolors=blue]{hyperref}  %
\usepackage{color}

\usepackage{amsmath,amssymb}  %
\usepackage{multirow}  %
\usepackage{listings}
\usepackage{enumitem}
\usepackage{xcolor}
\usepackage{subfig}
\usepackage{caption}
\usepackage{mathtools}
\usepackage{wrapfig}
\usepackage{colortbl}

\definecolor{artifactblue}{HTML}{0059A0}
\definecolor{parametergreen}{HTML}{539A55}
\definecolor{valuered}{HTML}{CF8369}
\definecolor{contextgrey}{HTML}{A0A0A0}
\definecolor{lightgrey}{HTML}{CFCFCF}

\colorlet{punct}{red!60!black}
\definecolor{background}{HTML}{F7F7F7}
\definecolor{delim}{RGB}{20,105,176}
\colorlet{cmnt}{magenta!60!black}

\lstdefinelanguage{plain}{
    stepnumber=1,
    numbersep=8pt,
    showstringspaces=false,
    breaklines=true,
    frame=lines,
    backgroundcolor=\color{background},
    morecomment=[s]{/*}{*/},
    commentstyle=\color{cmnt}\ttfamily,
    breakindent=0pt,
}
\lstdefinelanguage{json}{
    basicstyle=\normalfont\ttfamily,
    stepnumber=1,
    numbersep=8pt,
    showstringspaces=false,
    breaklines=true,
    frame=lines,
    backgroundcolor=\color{background},
    morecomment=[s]{/*}{*/},
    commentstyle=\color{cmnt}\ttfamily,
    literate=
     *{:}{{{\color{punct}{:}}}}{1}
      {,}{{{\color{punct}{,}}}}{1}
      {\{}{{{\color{delim}{\{}}}}{1}
      {\}}{{{\color{delim}{\}}}}}{1}
      {[}{{{\color{delim}{[}}}}{1}
      {]}{{{\color{delim}{]}}}}{1},
}

\begin{document}
\title{HyperPIE: Hyperparameter Information Extraction from Scientific Publications}
\titlerunning{HyperPIE: Hyperparameter IE from Papers}

\author{\href{https://orcid.org/0000-0001-5028-0109}{Tarek Saier}\inst{1} \and
\href{https://orcid.org/0000-0001-5354-5571}{Mayumi Ohta}\inst{2} \and
\href{https://orcid.org/0000-0001-6212-4779}{Takuto Asakura}\inst{3} \and
\href{https://orcid.org/0000-0001-5458-8645}{Michael F{\"a}rber}\inst{1}}
\authorrunning{T. Saier et al.}
\institute{Karlsruhe Institute of Technology, Karlsruhe, Germany
\email{\{tarek.saier,michael.faerber\}@kit.edu}\\ \and
Fraunhofer Institute for Systems and Innovation Research, Karlsruhe, Germany\\
\email{mayumi.ohta@isi.fraunhofer.de}\\ \and
The University of Tokyo, Tokyo, Japan\\
\email{takuto@is.s.u-tokyo.ac.jp}}

\maketitle              %
\begin{abstract}
Automatic extraction of information from publications is key to making scientific knowledge machine-readable at a large scale.
The extracted information can, for example, facilitate academic search, decision making, and knowledge graph construction.
An important type of information not covered by existing approaches is hyperparameters.
In this paper, we formalize and tackle hyperparameter information extraction (HyperPIE) as an entity recognition and relation extraction task.
We create a labeled data set covering publications from a variety of computer science disciplines.
Using this data set, %
we train and evaluate BERT-based fine-tuned models as well as five large language models: GPT-3.5, GALACTICA, Falcon, Vicuna, and WizardLM.
For fine-tuned models, we develop a relation extraction approach that achieves an improvement of 29\% $\text{F}_1$ over a state-of-the-art baseline.
For large language models, we develop an approach leveraging YAML output for structured data extraction, which achieves an average improvement of 5.5\% $\text{F}_1$ in entity recognition over using JSON.
With our best performing model we extract hyperparameter information from a large number of unannotated papers, and analyze patterns across disciplines.
All our data and source code is publicly available at 
\url{https://github.com/IllDepence/hyperpie}.

\keywords{Information Extraction, Scientific Text, Hyperparameter}
\end{abstract}

\section{Introduction}

\begin{figure}[bt]
  \centering
  \includegraphics[width=\linewidth]{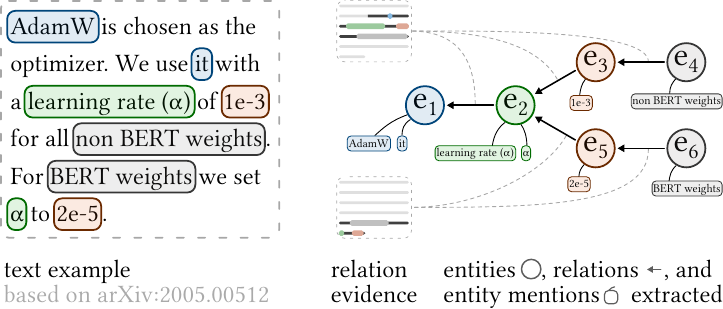}
  \caption{Illustration of hyperparameter information in a text example alongside the extracted entities and relations. Entity types are {\color{artifactblue}{research artifact}}, {\color{parametergreen}{parameter}}, {\color{valuered}{value}}, and {\color{contextgrey}{context}}. Relations are indicated by arrows.}
  \label{fig:schema-visual}
\end{figure}

Models capable of extracting fine-grained information from publications can make scientific knowledge machine-readable at a large scale.
Aggregated, such information can fuel platforms like Papers with Code\footnote{See \url{https://paperswithcode.com/}.} and the Open Research Knowledge Graph~\cite{orkg1,orkg2}, and thereby facilitate academic search, recommendation,  and reproducibility.
Accordingly, a variety of approaches for information extraction~(IE) from scientific text have been proposed~\cite{luan2018scierc,Jain2020scirex,semeval21_task8,semeval22_task12,Dunn2022}.  %

However, to the best of our knowledge, no approaches exist for the extraction of structured information on hyperparameter use from publications.
That is, information on \emph{with which parameters} researchers use methods and data. We refer to this information as ``hyperparameter information'' (see Fig.~\ref{fig:schema-visual}).
Hyperparameter information is important for several reasons. (1)~First, its existence in a paper is an indicator for reproducibility~\cite{Radd2019} and, when extracted automatically, can improve automated reproduction of results~\cite{sethi2018}. (2)~Second, in aggregate it can inform on both conventions in a field as well as trends over time. (3)~Lastly, it enables more fine-grained paper representations benefiting downstream applications based on document similarity, such as recommendation and search. %
Hyperparameter information is challenging to extract, because (1) it is usually reported in a dense format, (2) often includes special notation, and (3) operates on domain specific text (e.g. ``For Adam we set $\alpha$ and $\beta$ to 1e-3 and 0.9 respectively.'').

To address the lack of approaches for extracting this type of information, we define the task of ``hyperparameter information extraction'' (HyperPIE) and develop several approaches to it. Specifically, we formalize HyperPIE as an entity recognition (ER) and relation extraction (RE) task. We create a labeled data set spanning a variety of computer science disciplines from machine learning (ML) and related areas. The data set is created by manual annotation of paper full-texts, which is accelerated by a pre-annotation mechanism based on an external knowledge base. Using our data set, we train and evaluate both BERT-based~\cite{devlin2019} fine-tuned models as well as large language models (LLMs).
For the former, we develop a dedicated relation extraction model that achieves an improvement of 29\% $\text{F}_1$ compared to a state-of-the-art baseline.
For LLMs, we develop an approach leveraging YAML output for structured data extraction, which achieves a consistent improvement in entity recognition across all tested models, averaging at 5.5\% $\text{F}_1$.
Using our best performing model, we extract hyperparameter information from 
15,000 unannotated papers, and analyze patterns across ML disciplines of how authors report hyperparameters.
All our data and source code is made publicly available.\footnote{See \url{https://github.com/IllDepence/hyperpie}.}
In summary, we make the following contributions.

\begin{enumerate}
    \item We formalize a novel and relevant IE task (HyperPIE).
    \item We create a high quality, manually labeled data set from paper full-texts, enabling the development and study of approaches to the task.
    \item We develop two lines of approaches to HyperPIE and achieve performance improvements in both of them over solutions based on existing work.
    \item We demonstrate the utility of our approaches by application on large-scale, unannotated data, and analyze the extracted hyperparameter information.
\end{enumerate}

\section{Related Work}\label{sec:refwork}

\subsubsection{Fine-Tuned Models}

Named entity recognition (NER) and RE from publications in ML and related fields have been tackled by SciERC~\cite{luan2018scierc} and subsequently SciREX~\cite{Jain2020scirex}. The entity types considered are methods, tasks, data sets, and evaluation metrics. Proposed methods for the task utilize BiLSTMs, BERT and SciBERT~\cite{beltagy2019}. With both approaches, there is a partial overlap in entity types to our task, as we also extract methods and data sets. The key difference arises though the parameter and value entities we cover, which are a challenge in part due to their varied forms of notation (e.g. $\alpha$\,/\,alpha, or $0.001$\,/\,$1\times 10^{-3}$ / 1e-3).

IE models aiming to relate natural language to numerical values and mathematical symbols have been introduced at SemEval 2021 Task 8~\cite{semeval21_task8} and SemEval 2022 Task 12~\cite{semeval22_task12} respectively. Most of the proposed models base their processing of natural language on BERT or SciBERT. To handle numbers and symbols rendered in \LaTeX, as well as to accomplish RE between entity types with highly regular writing conventions (e.g. numbers and units such as ``5 ms''), rule-based approaches or dedicated smaller neural networks are commonly used.

Similarly, we find a level of regularity in how authors report parameters and values, and make use of that in our approach accordingly. In line with related work using fine-tuned models, we also use BERT and SciBERT for contextualized token embeddings.

\subsubsection{LLMs}

With the recent advances in LLMs, there has been a surge in efforts to utilize them for IE from scientific text. Nevertheless, their performance is not on par with dedicated models for NER and RE yet~\cite{Yang2023}.

An improtant concept for IE with LLMs is introduced by Agrawal et al.~\cite{Agrawal2022}: a ``resolver'' is a function that maps the potentially ambiguous output of an LLM to a defined, task specific output space. In their work, the authors extract singular values and lists from clinical notes using GPT-3. They use a variety of resolvers that perform steps like tokenization, removal of specific symbols or words, and pattern matching using regular expressions.

Work with similar output data complexity (values and lists) has also been done in the area of material science. Xie et al.~\cite{Xie2023} use GPT-3.5 to extract information on solar cells from paper full-text.
Similarly, Polak et al.~\cite{Polak2023} use ChatGPT %
to extract material, value, and unit information from sentences of material science papers. They define a conversational progression, in which they prompt the model generate tables, which are processed using simple string parsing rules.

An approach for IE of more complex information is proposed by Dunn et al.~\cite{Dunn2022}. They use GPT-3.5 to extract material information from materials chemistry papers. Given the hierarchical nature of the information to be extracted, the authors find simple output formats insufficient. To overcome this, they prompt the model to output the data in JSON format.%

Given hyperparameter information also is hierarchical (see Fig.~\ref{fig:schema-visual}), we adopt prompting LLMs to output data in a text based data serialization format. Different from the related work introduced above, we do not limit our experiments to API access based closed source LLMs, but also evaluate various open LLMs, because we recognize the importance of contributing efforts to the advancement of the more transparent, accountable, and reproducibility friendly side of this new and rapidly evolving area of research~\cite{Liesenfeld2023}.\\
\\
Besides IE from scientific publications, there have been efforts to extract hyperparameter schemata and constraints from Python docstrings~\cite{Baudart2020} using CNL grammars~\cite{Kuhn2014}, and from Python code~\cite{RakAmnouykit2021} using static analysis. Compared to our task setting, these rely on a known context (e.g. a \texttt{fit} method) and operate on constrained input (generated docstrings and source code instead).

\section{Hyperparameter Information Extraction}\label{sec:hyperpie}

\subsection{Task Definition}

We define HyperPIE as an ER+RE task with four entity classes ``research artifact'', ``parameter'', ``value'', and ``context'', and a single relation type. Briefly illustrated by a minimal example, in the sentence \textit{``During fine-tuning, we use the Adam optimizer with $\mathit{\alpha=10^{-4}}$.''}, the research artifact \textit{Adam} has the parameter $\mathit{\alpha}$ which is set to the value $\mathit{10^{-4}}$ in the context \textit{During fine-tuning}.

The entity classes are characterized as follows. A ``research artifact'', within the scope of our task, is an entity used for a specific purpose with a set of variable aspects that can be chosen by the user. These include methods, models, and data sets.\footnote{Broader definitions in other contexts also include software in general, empirical laws, and ideas~\cite{Lin2022}. For our purposes, however, above specific definition is more useful.} A ``parameter'' is a variable aspect of an artifact. This includes model parameters, but also, for example, the size of a sub-sample of a data set. A ``value'' expresses a numerical quantity and in our task is treated like an entity rather than a literal. Lastly, a ``context'' can be attached to a value if the value is only valid in that specific context. The single relation type relates entities as follows: parameter\,$\rightarrow$\,research artifact, value\,$\rightarrow$\,parameter, and context\,$\rightarrow$\,value.
Co-reference relations %
implicitly exist between the mentions of a common entity (e.g. ``AdamW'' and ``it'' in Fig.~\ref{fig:schema-visual}). That is, if an entity has multiple mentions within the text, they are considered co-references to each other.

The scope of the IE task comprises the extraction of entities, their relations, and the identification of all their mentions in the text (and thereby implicitly co-references). %
Furthermore, we specifically consider IE from text, and not from tables, graphs, or source code.\footnote{We leave investigating multi-modal IE pipelines (text/code/graphs) for future work.}

\subsection{Data Set Construction}\label{sec:data-set-contruction}

Because HyperPIE is a novel task, we cannot rely on existing data sets for training and evaluating our approaches. We therefore create a new data set by manually annotating papers. As our data source we chose unarXive~\cite{Saier2023unarXive}, because it includes paper full-texts and, most importantly, retains mathematical notation as \LaTeX. This is crucial because parsing such notation from PDFs is prone to noise, which would be problematic for our parameter and value entities.  %

To ensure we cover a wide variety of artifacts and discipline specific writing conventions, we use papers from multiple ML related fields. Specifically, these are Machine Learning (ML), Computation and Language (CL), Computer Vision (CV), and Digital Libraries (DL), which make up 143,203 papers in unarXive.\footnote{The respective arXiv categories are cs.LG, cs.CL, cs.AI, and cs.DL. See \url{https://arxiv.org/category_taxonomy} for a more detailed description.}

We base our annotation guidelines on the widely used ACL RD-TEC guideline\footnote{See \url{http://pars.ie/publications/papers/pre-prints/acl-rd-tec-guidelines-ver2.pdf}.}~\cite{Qasemizadeh2016}. To make sure our resulting annotations are able to properly capture how authors report hyperparameters in text, we perform two annotation rounds: (1) an initial exploratory round, the results of which are used to refine the annotation guidelines and inform later model development, and (2) the main annotation round, the results of which constitute our data set used for model training and evaluation. In the following, both steps are described in more detail.

\subsubsection{Initial Annotation Round}\label{sec:exploreannot}

We heuristically pre-filter our ML paper corpus for sections reporting on hyperparameters.\footnote{We filter based on key phrases (``use'', ``set'', etc.), numbers, and \LaTeX\ math content.} Annotators then inspect these sections, select a continuous segment of text that contain hyperparameter information, and make their annotations. This task is performed independently by two annotators and results in a total of 151 text segments (131 unique, $2\times10$ annotated by both). The annotated text segments contain 1,345 entities and 1,110 relations.

\begin{figure}
    \centering
    \subfloat[text segments\label{fig:text-seg-size}]{%
        \includegraphics[width=.35\linewidth]{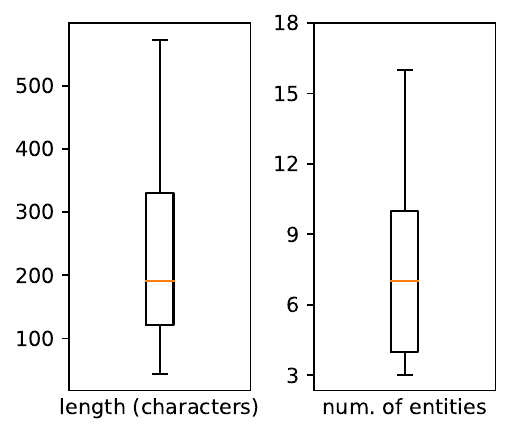}
    }
    \subfloat[relation distances (\#chars)\label{fig:rel-dists}]{%
        \includegraphics[width=.53\linewidth]{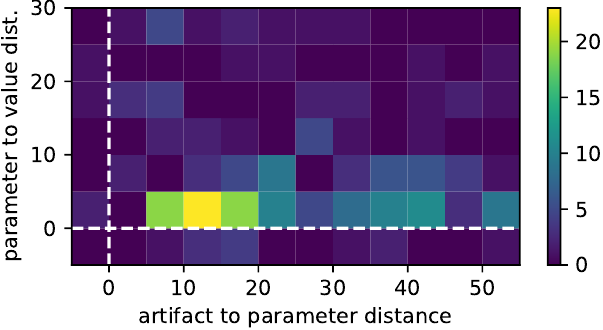}
    }
    \caption{Observations of initial annotation round}
    \label{fig:init-annot}
\end{figure}

As shown in Figure~\ref{fig:text-seg-size}, we observe text segments reporting on hyperparameters to generally have a length below 600 characters. %
We furthermore see that most text segments contain between 3 and 15 entities. %
Lastly, in Figure~\ref{fig:rel-dists}, show distances between artifacts and their parameters, as well as parameters and their values. We see that artifacts usually are mentioned before their parameters (78\%), and parameters before their values (93\%). The reverse cases also exists, but are less common.
Additionally, we can see that values are most commonly reported right after their parameter, while there is a higher variability in distances between parameters and artifacts.
Based on above observations %
we determine the unit of annotation for the final round to be one paragraph (on average 563.4 characters long in our corpus), as it is sufficient to capture hyperparameters being reported.

The inter annotator agreement (IAA, reported as Cohen's kappa) of the text segments annotated by both annotators is 0.867 for entities and 0.737 for relations\footnote{Measured by the character level entity class and character level relation target span agreement respectively.} (strong to almost perfect agreement) which is compares favorably to SciERC~\cite{luan2018scierc} with an IAA of 0.769 for entities and 0.678 for relations.

\subsubsection{Main Annotation Round}

In our main annotation round we annotate whole papers (paragraph by paragraph) instead of pre-filtered text-segments. This is done to ensure that the final annotation result reflects data as it will be encountered by a model during inference---that is, containing a realistic amount of paragraphs that have no information on hyperparameters, or, for example, only mention research artifacts but no parameters.

Similar to related work~\cite{Jain2020scirex}, we use Papers with Code as an external knowledge base to pre-annotate entity candidates to make the annotation process more efficient. In a similar fashion, we use annotator's previously annotated entity mentions for pre-annotation. Pre-annotated text spans are, as the name suggests, set automatically, but need to be checked by annotators manually.

Through this process we annotate 444 paragraphs, which contain 1,971 entities and 614 relations. The entity class distribution is 1,134 research artifacts, 131 parameters, 662 values, and 44 contexts. The annotation data is provided in a JSON structure as shown in Figure~\ref{fig:schema-visual}, as well as in the W3C Web Annotation Data Model\footnote{See \url{https://www.w3.org/TR/annotation-model/}.} to facilitate easy re-use and compatibility with existing systems.

\section{Methods}\label{sec:methods}

We approach hyperparameter information extraction in two ways. First, we build upon established ER+RE methods and develop an approach using a fine-tuned model in a supervised learning setting. Second, given the recent promising advances with LLMs, we develop an approach utilizing LLMs in a zero-shot and few-shot setting.

\subsection{Fine-Tuned Models}

We base our fine-tuned model approach on PL-Marker~\cite{Ye2022}, the currently best performing model on SciERC. Specifically, we use the ER component of PL-Marker. Our reason is that (1)~the text our model will be applied on is of the same type as in SciERC (ML publications), and (2)~there is some correspondence between the entities to be identified---namely our entity class ``research artifact'' including methods and datasets, which are both entity classes in SciERC.

For RE we develop an approach that utilizes token embeddings as well as relative entity distance and entity class pairings. This is motivated by the fact that (a) we observed a high level or regularity in the relative distance of research artifact, parameter, and value mentions\footnote{We note that these observations where made during the initial exploratory annotation round (Sec.~\ref{sec:exploreannot}) and not during annotation of the evaluation data.} (see Fig.~\ref{fig:init-annot}), and (b) relations only exist between specific pairs of entity types.

\begin{figure}[bt]
  \centering
  \includegraphics[width=.8\linewidth]{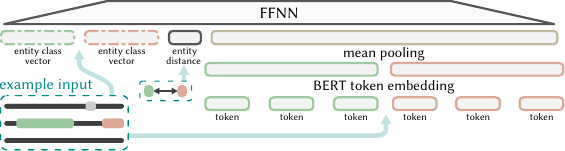}
  \caption{RE with emphasis on entity candidate pair types and distance.}
  \label{fig:ffnn-re-sub-visual}
\end{figure}

In Figure~\ref{fig:ffnn-re-sub-visual} we show a schematic depiction of our new relation extraction component. Entity candidate pair classes as well as the relative distance between the entities in the text are used as a dedicated model input, BERT token embeddings of the entity mentions are combined using mean pooling. These inputs are fed into a feed-forward neural network FFNN for prediction. Formally, the model performs pairwise binary classification as $\text{FFNN}(E^c_0, E^c_1, E^d, E^T)$, where $E^c_i$ are class vectors, $E^d$ encodes candidate distance, and $E^T$ is the token pair embedding calculated as $E^T=\frac{1}{|T|}\sum^{|T|}_{i=0}\text{BERT}(t_i)$, the mean of the pair's tokens $t_i\in|T|$.

During the development of our model we also experiment with concatenation in favor of mean pooling to preserve information on the order of the entities, but find that mean pooling results in better performance. Furthermore, we investigated the use of SciBERT instead of BERT, but find that regular BERT embeddings give us better results, despite our model handling scientific text.

\subsection{LLM}\label{sec:methLLMs}

We develop our LLM approach for a zero-shot and a few-shot setting. This means the models perform the IE task based on either instructions only (zero-shot), or instructions and a small number of examples (few-shot).

Performing IE using LLMs in zero-shot or few-shot settings requires the desired structure of the output data to be specified within the model input. In simple cases (e.g. numbers or yes/no decisions) this can be achieved by an in-line specification of the format in natural language (e.g. ``The answer (arabic numerals) is'')~\cite{Kojima2022}. IE from scientific publications, however, often seeks to extract more complex information. To achieve this, the model can be tasked to produce output in a text based data serialization format such as JSON, as done in previous work~\cite{Dunn2022}. Especially for complex structured predictions, few-shot prompting has been shown to further boost in-context learning (ICL) accuracy and consistency at inference time~\cite{Brown2020}.

Drawing from techniques used in previous work approaching other IE tasks, we investigate several prompting strategies to build our approach.

\begin{enumerate}
    \item \textit{Multi-stage prompting}~\cite{Polak2023}: first determine the presence of hyperparameters information; if present, extract the list of entities; lastly, determine relations.
    \item \textit{In-text annotation}~\cite{Wang2023}: let the input text be repeated with entity annotations, e.g. repeat ``We use BERT for ...'' as ``We use [a1|BERT] for ...''. %
    \item \textit{Data serialization format}~\cite{Dunn2022}: specify a serialization format in the promt that is parsed afterwards; then match in-text mentions in the input.
    \item \textit{(3)+(2)}: prompt as in (3); then match in-text mentions using (2).
\end{enumerate}

We find (1) to lead to problems with errors propagation along steps. With (2) and (4) we frequently see alterations in the reproduced text. Accordingly, we use prompt type (3) for our approach---specifying a data serialization format in the prompt.
While existing work uses the JSON format for this~\cite{Dunn2022}, we use YAML, as it is less prone to ``delimiter collision'' problems due to its minimal requirements for structural characters.\footnote{See \url{https://yaml.org/spec/1.2.2/}.} 
In doing so, we expect to avoid problems with LLM output not being parsable. Our overall LLM approach looks as follows.

\subsubsection{Zero-shot} We build our zero-shot prompts from the following consecutive components: \texttt{[task]\allowbreak[input text]\allowbreak[format]\allowbreak[completion prefix]}.
In \texttt{[task]} we specify the information to extract, i.e. research artifacts, their parameters, etc. \texttt{[input text]} is the paragraph from which to extract the information. \texttt{[format]} defines the output YAML schema.\footnote{Examples can be found at \url{https://github.com/IllDepence/hyperpie}.} \texttt{[completion prefix]} is a piece of text that directly precedes the LLM's output, such as \textit{``ASSISTANT:~''}.
To generate predictions based on LLM output, we pass it to a standard YAML parser after cleansing (e.g. removing text around the YAML block).
For each used LLM model, we individually perform prompt tuning. Here we determine, for example, if a model gives better results when the \texttt{[completion prefix]} includes the beginning of the serialized output (e.g., ``\texttt{-{}-{}-\allowbreak\textbackslash n\allowbreak text\_\allowbreak contains\_\allowbreak entities:}'') or if this leads to a deterioration in output quality.

\subsubsection{Few-shot} Our few-shot approach makes the following adjustments to the method described above. Prompts additionally include a component \texttt{[examples]}, which are valid input output pairs sampled by their cosine similarity to the input text. Specifically, for an input text from a document X, we sample the five most similar paragraphs from all ground truth documents excluding X. As these examples can be confused with the input text, we re-position the input text to appear \emph{after} the examples. The resulting prompt structure we use for our few-shot approach is as follows: \texttt{[task]\allowbreak[format]\allowbreak[examples]\allowbreak[input text]\allowbreak[completion prefix]}.

LLMs reaching a sufficient context size for a few-shot approach to our task are a recent development. We can therefore additionally make use of other recently added capabilities. Specifically, we make use of generation constrains via a gBNF grammar\footnote{See \url{https://github.com/ggerganov/llama.cpp/pull/1773}.} to enforce LLM output according to our data scheme, allowing us to mitigate parsing errors.

\section{Experiments}\label{sec:experiments}

We evaluate the fine-tuned models and LLM approach against baselines from existing work. Both evaluations are performed on our data set described in Section~\ref{sec:data-set-contruction}. Metrics used to measure prediction performance are precision, recall and $\text{F}_1$ score, abbreviated as P, R and $\text{F}_1$ respectively.

\subsection{Fine-Tuned Models}

We use PL-Marker, the currently best performing model on SciERC, as our baseline.
Models are trained and evaluated using 5-fold cross validation (3 folds training, 1 dev, 1 test).
We train the ER component of PL-Marker as done in~\cite{Ye2022}, using \textit{scibert-scivocab-uncased} as the encoder, Adam as the optimizer, a learning rate of 2e-5, and 50 training epochs. Regarding the two RE components we compare, the PL-Marker RE component is trained using \textit{bert-base-uncased}, Adam, a learning rate of 2e-5, and 10 training epochs. Our own RE component also uses \textit{bert-base-uncased}, Adam as the optimizer, and is trained with a learning rate of 1e-3 for 90 epochs.\footnote{The two RE models we compare require different learning rates and number of training epochs, because their architecture varies significantly.}
The models are trained and evaluated on a server with a single GeForce~RTX~3090 (24\,GB).

\subsubsection{Results}

\begin{figure}[tb]
  \centering
  \includegraphics[width=.8\linewidth]{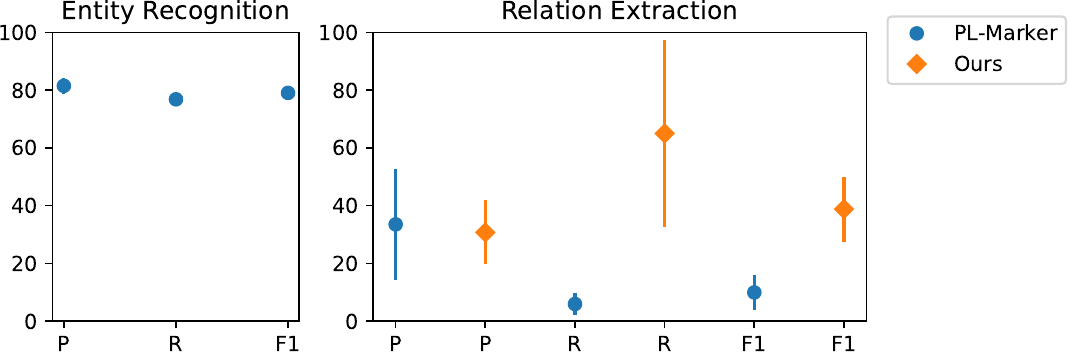}
  \caption{Fine-tuned model evaluation (5-fold cross validation).}
  \label{fig:finetunedeval}
\end{figure}

In Figure~\ref{fig:finetunedeval} we show the results of PL-Marker ER (used for both models) as well as the PL-Marker RE component and our RE model. For ER we evaluate exact matches (no partial token overlap). In the case of RE, each entity pair is predicted as having a relation or not---as there is just one relation type.

Mean ER performance is 81.5, 76.8, and 79.0 (P, R, $\text{F}_1$). For RE, the precision of PL-Marker and our model are similar at 33.5 and 30.7 respectively, but our model performs more consistent. PL-Marker only achieves a very low recall of 5.9, whereas our model, while showing large variability, achieves a mean of 65.0. The resulting $\text{F}_1$ scores are 9.9 for PL-Marker and 38.8 for our model.

\subsubsection{Analysis}

\begin{wraptable}[8]{r}{4cm}\vspace*{-2\baselineskip}%
  \caption{Ablation study}
  \label{tab:finetunedablation}
  \begin{tabular}{lccc}
    \hline
    Used & P [\%] & R [\%] & $\text{F}_1$\,[\%] \\
    \hline
    \textvisiblespace CD & 15.5 & 8.8 & 11.1 \\
    T\textvisiblespace D & 16.6 & 29.8 & 19.6 \\
    TC\textvisiblespace & 26.5 & 65.0 & 35.5 \\
    TCD & \textbf{30.7} & \textbf{65.0} & \textbf{38.8} \\
    \hline
  \end{tabular}
\end{wraptable}

Token level ER performance across entity classes (none, artifact, parameter, value, context) is at 98.5\%, 77.8\%, 47.9\%, 84.4\%, 0\% $\text{F}_1$. That is, the model does not predict contexts and struggles with parameters, but artifacts and values are predicted reliably. For our RE model, we observe that value-parameter relations are more reliably predicted than parameter-artifact relations.

\begin{wrapfigure}[13]{r}{3cm}\vspace*{-2\baselineskip}%
  \includegraphics[width=3cm]{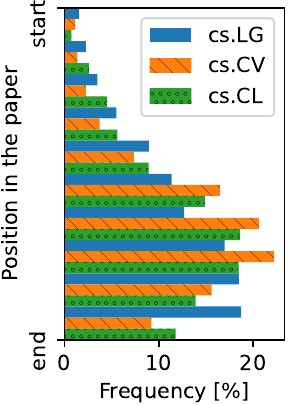}
  \vspace{-0.6cm}
  \caption{Mentioning position in papers.}
  \label{fig:hyperparam_info_pos}
\end{wrapfigure}

To assess the impact of the different components in our RE model, we perform an ablation study with the same 5-fold cross-validation setup as above. In Table~\ref{tab:finetunedablation}, showing its results, we can see that removing the BERT token embeddings (T) results in the largest performance loss, followed by entity class embeddings (C) and entity distance (D). Removing any of the inputs results in worse predictions.

Finally, we apply our full model to a random sample of 15,000 papers. Analyzing the results, we find hyperparameters (artifact, parameter, value triples) are reported in 36\% of ML papers, 42\% of CV papers, 36\% of CL papers, and 7\% of DL papers. In Figure~\ref{fig:hyperparam_info_pos} we further look at the distribution of the information across the length of papers (excluding DL as not being representative). We can see a clear tendency towards the latter half of papers.

\subsection{LLMs}

\begin{table}[tb]
\centering
  \caption{LLM selection (size in number of parameters).}
  \label{tab:llmselection}
  \begin{tabular}{llc}
    \hline
    Model & Variant & Size \\
    \hline
    WizardLM~\cite{xu2023wizardlm2023}
    & \texttt{WizardLM-13B-V1.1} & 13\,B \\
    Vicuna${}_{4k}$~\cite{vicuna2023}
    & \texttt{vicuna-13b-v1.3} & 13\,B \\
    Vicuna${}_{16k}$~\cite{vicuna2023}
    & \texttt{vicuna-13b-v1.5-16k} & 13\,B \\
    Falcon~\cite{falcon40b-huggingface}
    & \texttt{falcon-40b-instruct} & 40\,B \\
    GALACTICA~\cite{GALACTICA2022}
    & \texttt{galactica-120b} & 120\,B \\
    GPT-3.5~\cite{Brown2020gpt3}
    & \texttt{text-davinci-003} & 175\,B \\
    \hline
    \end{tabular}
\end{table}

For our LLM experiments we chose a variety of models, with sizes ranging from 13\,B to 175\,B parameters, as shown in Table~\ref{tab:llmselection}.
We chose WizardLM~\cite{xu2023wizardlm2023} as it is meant to handle complex instructions, 
Vicuna~\cite{vicuna2023} due to its performance relative to its size, 
Falcon~\cite{falcon40b-huggingface} because of its alleged performance, 
and GALACTICA~\cite{GALACTICA2022} because it was trained on scientific text.
Vicuna${}_{16k}$ is a model extended using Position Interpolation~\cite{chen2023} based on Rotary Positional Embeddings~\cite{su2021}, which makes it the only model in our experiments with a sufficient context size for a few-shot evaluation.

The models are run as follows. GPT-3.5 is accessed through its official API.
All open models are run on a high performance compute cluster.
Vicuna${}_{4k}$ and WizardLM are run on nodes with $4\times$NVIDIA Tesla V100 (32\,GB).
GALACTICA, Falcon, and Vicuna${}_{16k}$ are run on nodes with $4\times$NVIDIA A100 (80\,GB).

As a baseline, we use a JSON variant for each model, where the \texttt{[format]} and \texttt{[examples]} compontents of prompts use JSON, and compare it to the respective YAML version. All models are used with greedy decoding (temperature = 0) for the sake of reproducibility.

\subsubsection{Results}

\begin{table}[tb]
  \centering
  \caption{Prediction performance of LLM models. Subscripts (${}_{\Delta\pm n}$) show the delta in $\text{F}_1$ from JSON to YAML output of each model. Format: \textbf{best}, \underline{second}.}
  \label{tab:llmeval}
  \begin{tabular}{ll|ccc|ccc}
    \hline
    \multicolumn{2}{l|}{\textbf{Zero-shot}} &
    \multicolumn{3}{c|}{Entity Recognition} &
    \multicolumn{3}{c}{Relation Extraction} \\
    \hline
    Model & Output & P [\%] & R [\%] & $\text{F}_1$\,[\%] &
                     P [\%] & R [\%] & $\text{F}_1$\,[\%] \\
    \hline

    \arrayrulecolor{lightgrey}\cline{1-2}\arrayrulecolor{black}
    \multirow{2}{*}{WizardLM} &
    JSON & 6.9 & 11.3 & 8.6
              & 0.1 & 0.8 & 0.1 \\
    \ & YAML & 9.7 & 35.6 &
    \hphantom{${}_{\Delta\text{+6.7}}$}
    15.3{\color{parametergreen}{${}_{\Delta\text{+6.7}}$}}
              & 0.1 & 1.5 &
    \hphantom{${}_{\Delta\text{+0.0}}$}
    0.1{\color{contextgrey}{${}_{\Delta\text{+0.0}}$}}  \\
    \arrayrulecolor{lightgrey}\cline{1-2}\arrayrulecolor{black}

    \multirow{2}{*}{Vicuna${}_{4k}$} &
    JSON & 15.1 & 9.3 & 11.5
              & 0.7 & 3.8 & 1.2 \\
    \ & YAML & 17.3 & 31.5 &
    \hphantom{${}_{\Delta\text{+10.8}}$}
    22.3{\color{parametergreen}{${}_{\Delta\text{+10.8}}$}}
              & 0.0 & 0.8 &
    \hphantom{${}_{\Delta\text{-1.1}}$}
    0.1{\color{valuered}{${}_{\Delta\text{-1.1}}$}}  \\
    \arrayrulecolor{lightgrey}\cline{1-2}\arrayrulecolor{black}

    \multirow{2}{*}{Falcon} &
    JSON & \textbf{37.1} & 5.9 & 10.2
              & 0.0 & 0.0 & 0.0 \\
    \ & YAML & 32.7 & 14.2 &
    \hphantom{${}_{\Delta\text{+9.6}}$}
    19.8{\color{parametergreen}{${}_{\Delta\text{+9.6}}$}}
              & 0.0 & 0.0 &
    \hphantom{${}_{\Delta\text{+0.0}}$}
    0.0{\color{contextgrey}{${}_{\Delta\text{+0.0}}$}}  \\
    \arrayrulecolor{lightgrey}\cline{1-2}\arrayrulecolor{black}

    \multirow{2}{*}{GALACTICA} &
    JSON & 25.9 & 15.7 & 19.5
              & 0.1 & 2.3 & 0.3 \\
    \ & YAML & 23.1 & 19.5 &
    \hphantom{${}_{\Delta\text{+1.6}}$}
    21.1{\color{parametergreen}{${}_{\Delta\text{+1.6}}$}}
              & 0.0 & 0.8 &
    \hphantom{${}_{\Delta\text{-0.2}}$}
    0.1{\color{valuered}{${}_{\Delta\text{-0.2}}$}}  \\
    \arrayrulecolor{lightgrey}\cline{1-2}\arrayrulecolor{black}

    \multirow{2}{*}{GPT-3.5} &
    JSON & 27.9 & \textbf{42.8} & \underline{33.8}
              & \underline{5.4} & \underline{10.7} & \underline{7.2} \\
    \ & YAML & \underline{34.0} & \underline{41.7} &
    \hphantom{${}_{\Delta\text{+3.6}}$}
    \textbf{37.4}{\color{parametergreen}{${}_{\Delta\text{+3.6}}$}}
              & \textbf{5.8} & \textbf{12.2} &
    \hphantom{${}_{\Delta\text{+0.6}}$}
    \textbf{7.8}{\color{parametergreen}{${}_{\Delta\text{+0.6}}$}}  \\

  \hline
    \multicolumn{2}{l|}{\textbf{5-shot}} &
    \multicolumn{3}{c|}{Entity Recognition} &
    \multicolumn{3}{c}{Relation Extraction} \\
  \hline

    \multirow{2}{*}{Vicuna${}_{16k}$} &
    JSON & \underline{34.4} & \underline{46.7} & \underline{39.6}
              & \underline{0.8} & \underline{4.6} & \underline{1.3} \\
    \ & YAML & \textbf{43.9} & \textbf{44.1} &
    \hphantom{${}_{\Delta\text{+0.4}}$}
    \textbf{44.0}{\color{parametergreen}{${}_{\Delta\text{+0.4}}$}}
              & \textbf{4.5} & \textbf{9.9} &
    \hphantom{${}_{\Delta\text{+4.8}}$}
    \textbf{6.1}{\color{parametergreen}{${}_{\Delta\text{+4.8}}$}}  \\
  \hline
  \end{tabular}
\end{table}

In Table~\ref{tab:llmeval}, show the prediction performance of all models and prompt variants. Overall, LLM performance does not reach the level of our pre-trained models. For zero-shot, we observe the best performance with both GPT-3.5 variants, where YAML outperforms JSON (+3.6\% ER and +0.6\% RE in $\text{F}_1$ score). The second highest ER $\text{F}_1$ score by model is achieved by Vicuna${}_{4k}$ (22.3), despite its size being less than a 10th that of GPT-3.5. For RE, however, even the best model only reaches 7.8\%. With our few-shot approach, we are able to considerably improve performance between Vicuna models (+27\% ER and +6\% RE in $\text{F}_1$), surpassing the zero-shot performance of GPT-3.5 in ER.
Lastly, we see that using YAML leads to better ER results accross all six models, with ER performance being comparable or improved as well.

\subsubsection{Analysis}

\begin{figure}[tb]
  \centering
  \includegraphics[width=\linewidth]{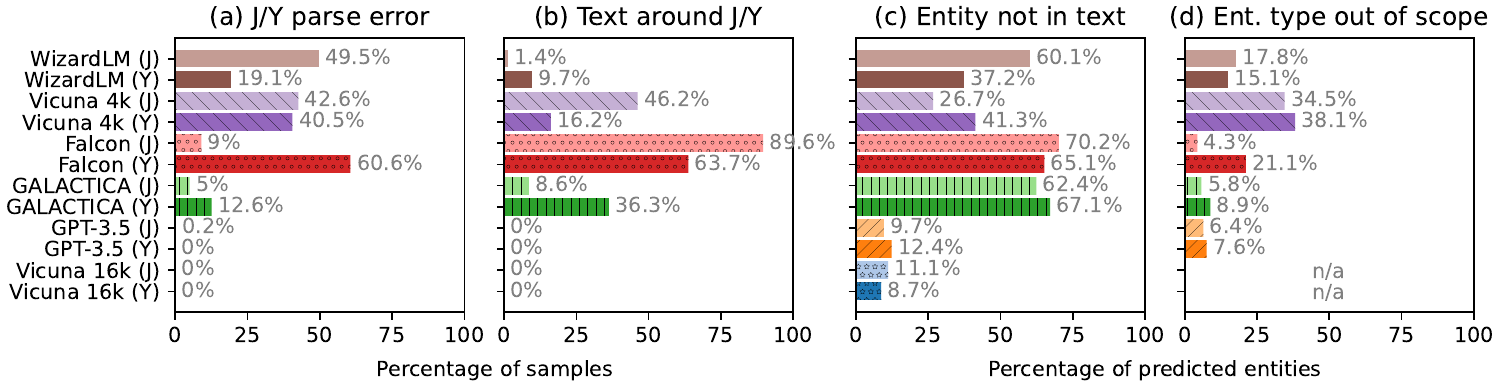}
  \caption{Parsing success, format adherence, hallucinations, and scope adherence of LLM generated JSON (J) and YAML (Y).}
  \label{fig:yamlVSjson}
\end{figure}

In Figure~\ref{fig:yamlVSjson} we show an analysis of the steps leading up to model prediction. Focussing first on the zero-shot models (upper five) we observe the following across the four plots from left to right. (a)~For three of five models, prompting for YAML leads to fewer parsing errors. (b)~Unwanted text around the extracted data is generated more/less by two models each. (c)~Hallucinated entities and (d)~out of scope entities appear overall slightly more often for in YAML compared to JSON. For our few-shot approach (bottom model), we see that the use of a grammar (a, b)~prevents all output format issues. Furthermore (c)~hallucinated entities are reduced. (d)~Out of scope entities can not be evaluated, because our in-context examples lead to frequent omission of type information in the output.

Through manual analysis we identify a common cause for parsing errors in JSON output to be boolean values (e.g. for ``\texttt{text\_\allowbreak contains\_\allowbreak entities:}'') being copied by the LLM as ``true/false'' from the promt. We furthermore find that ``entities not in the text'' can arise from unsolicited \LaTeX\ parsing by the LLM (e.g. ``\verb|\lambda|'' in text $\rightarrow$ ``$\lambda$'' in YAML). Prompting for \emph{verbatim} parameter/value strings does not mitigate this.

\section{Discussion}\label{sec:discussion}

Our overall results, with a top performance of 79\% $\text{F}_1$ for entity recognition and 39\% $\text{F}_1$ for relation extraction, show that extraction of hyperparameter information from scientific text can be accomplished to a degree that yields sound results. There are, however, challenges that remain, such as more reliable entity recognition of parameters and contexts, as well as more reliable relation extraction in general. Our novel data set enables further development of approaches from hereon. Our IE results on large-scale unannotated data give an indication of possible downstream analyses and applications. Here we see large potential for reproducibility research, faceted search, and recommendation.

Our LLM evaluation shows that for IE tasks dealing with complex information, the choice of text based data serialization format can have a considerable impact on performance, even when using grammar based generation constrains. Additionally, we can see that in-context learning enabled by larger context sizes, as well as grammars, are an effective method to improve IE performance.

\subsubsection{Limitations}

(1)~Our work considers HyperPIE from text. This is sensible for a focussed approach, but downstream applications could furthermore benefit from composite pipelines also targeting extraction from tables, source code, etc.
(2)~We do not test transferability of methods to domains outside of ML related fields. It would require domain expertise to find useful definitions for hyperparameters in each respective domain.
(3)~Our LLM evaluation does not cover fine-tuning. Presupposing the existence of a large enough training data set, this would be a valuable addition the overall investigation.
(4)~Defining our YAML/JSON output format hierarchically means that only values associated with parameters and parameters associated with artifacts can be extracted.
(5)~Lastly, our data and experiments unfortunately are limited to English text only and do not cover other languages.

\section{Conclusion}\label{sec:conclusion}

We formalize the novel ER+RE task HyperPIE and develop approaches for it, thereby expanding IE from scientific text to hyperparameter information. To this end, we create a manually labeled data set spanning various ML fields. In a supervised learning setting, we propose a BERT-based model that achieves an improvement of 29\% $\text{F}_1$ in RE compared to a state-of-the-art baseline. Using the model, we perform IE on a large amount of unannotated papers, and analyze patterns of hyperparameter reporting across ML disciplines. In a zero-/few-shot setting, we propose an LLM based approach using YAML for complex IE, achieving an average improvement of 5.5\% $\text{F}_1$ in ER over using JSON. We furthermore achieve large performance gains for LLMs using grammar based generation constrains and in-context learning. In future work, we plan to investigate fine-tuning LLMs, as well as additional practical use cases for data extracted from large publication corpora, such as knowledge graph construction.

\section*{Author Contributions}  %
Tarek Saier: Conceptualization, Data curation, Formal analysis, Methodology, Software, Visualization, Writing -- original draft, Writing -- review \& editing. Mayumi Ohta: Conceptualization (LLM few-shot), Formal analysis (LLM few-shot), Methodology (LLM few-shot), Software (LLM few-shot), Writing -- original draft (support). Takuto Asakura: Conceptualization, Writing -- review \& editing. Michael F{\"a}rber: Writing -- review \& editing.

\section*{Acknowledgements}
This work was partially supported by the German Federal Ministry of Education and Research (BMBF) via [KOM,BI], a Software Campus project (01IS17042).
The authors acknowledge support by the state of Baden-W{\"u}rttemberg through bwHPC.
We thank Nicholas Popovic for extensive feedback on the experiment design and prompt engineering. We thank Tarek Gaddour for feedback during the annotation scheme development, and Xiao Ning for input during early model development.

\bibliographystyle{splncs04}
\bibliography{paper}

\end{document}